\documentclass[10pt,twocolumn,letterpaper]{article}
\usepackage{array}

\usepackage{cvpr}
\usepackage{times}
\usepackage{epsfig}
\usepackage{graphicx}
\usepackage{amsmath}
\usepackage{amssymb}
\usepackage[utf8]{inputenc}
\usepackage[T1]{fontenc}

\usepackage[pagebackref=true,breaklinks=true,letterpaper=true,colorlinks,bookmarks=false]{hyperref}

\cvprfinalcopy 

\def\cvprPaperID{****} 

\ifcvprfinal\fi
\begin{document}

\title{Project Report : CS 7643}

\author{John Christopher Tidwell\\
Georgia Institute of Technology\\
CS-7643-O01\\
{\tt\small jtidwell8@gatech.edu}
\and
John Storm Tidwell\\
Georgia Institute of Technology\\
CS-7643-O01\\
{\tt\small jtidwell9@gatech.edu}
}

\maketitle

\begin{abstract}
This project addresses the challenge of automated stock trading, where traditional methods and direct reinforcement learning (RL) struggle with market noise, complexity, and generalization. Our proposed solution is an integrated deep learning framework combining a Convolutional Neural Network (CNN) to identify patterns in technical indicators formatted as images, a Long Short-Term Memory (LSTM) network to capture temporal dependencies across both price history and technical indicators, and a Deep Q-Network (DQN) agent which learns the optimal trading policy (buy, sell, hold) based on the features extracted by the CNN and LSTM. The CNN and LSTM act as sophisticated feature extractors, feeding processed information to the DQN, which learns the optimal trading policy (buy, sell, hold) through RL. We trained and evaluated this model on historical daily stock data, using distinct periods for training, testing, and validation. Performance was assessed by comparing the agent's returns and risk on out-of-sample test data against baseline strategies, including passive buy-and-hold approaches. This analysis, along with insights gained from explainability techniques into the agent's decision-making process, aimed to demonstrate the effectiveness of combining specialized deep learning architectures, document challenges encountered, and potentially uncover learned market insights.
\end{abstract}

\section{Introduction/Background/Motivation}

We aimed to create an automated system that decides when to buy, sell, or hold stocks to achieve better financial returns compared to a passive buy-and-hold strategy while simultaneously seeking to manage risk through learned behavior. The core problem this project addressed was making sequential, effective trading decisions based on complex, high-variance historical market data. Our goal was to develop a deep learning model that learns an optimal trading policy through interaction with historical data.

Currently, stock trading decisions are often made by human experts using technical analysis or automated systems following predefined rules based on market indicators. More advanced methods utilize machine learning for price prediction or apply reinforcement learning (RL) agents that learn through trial and error [1]. However, these approaches face significant limitations. Financial markets exhibit nonstationary behavior and significant noise, making it difficult for static rule-based systems or simple prediction models to adapt. Applying RL directly to raw market data is challenging due to the high-dimensional state space and the sparse, delayed nature of profit-based rewards, which can lead to models overfitting historical patterns that fail to generalize.

A successful automated trading agent offers potential benefits to a wide range of users, including individual investors, professional traders, and financial institutions managing large portfolios. By potentially capturing more gains during market uptrends and reducing losses during downturns, such a system could lead to more stable and higher long-term investment growth. Avoiding significant losses, especially during market crashes, is crucial to preserving capital. Ultimately, better trading decisions translate to improved financial outcomes, providing a strong incentive to develop more effective algorithmic trading strategies.

We used historical daily stock market data for 16 specific companies: Xerox (XRX), International Business Machines (IBM), and HP Inc. (HP), Bristol Myers Squibb (BMY), Pfizer (PFE), UnitedHealth Group (UNH), Eli Lilly and Company (LLY), JPMorgan Chase \& Co. (JPM), American Express (AXP), American International Group (AIG), Nike (NKE), PepsiCo (PEP), The Home Depot (HD), General Electric (GE), Caterpillar Inc. (CAT) and Honeywell International (HON). For each company and day, we used open, high, low, closed, and adjusted closed prices. We also calculated seven common technical indicators: Relative Strength Index (RSI), Momentum, Percentage Price Oscillator (PPO), Stochastic Oscillator (\% K), Bollinger Bands (\% B), Fibonacci Retracement levels, and Moving Average Convergence Divergence (MACD). This resulted in 12 features per day (5 price types + 7 indicators). Based on the configuration parameters, this data was formatted into 9x9 "images". Each image had 12 channels corresponding to the 12 features, providing a structured input suitable for convolutional processing. This 9x9 spatial layout was a design choice for the CNN input dimensions. The 12 channels also served as the sequence length for the LSTM component. We used distinct time periods for training (Jan 1, 2000 - Dec 31, 2005), testing (Jan 1, 2008 - Dec 31, 2016), and validation (Jan 1 2016 - Dec 31, 2016). The complete dataset can be found here \url{https://github.com/jstidwell/tidwell_gatech_spring_2025_cs7643} Using separate, chronologically ordered periods, including different market conditions like the 2008 financial crisis, helps ensure our model is evaluated realistically on its ability to generalize to unseen future data without any lookahead bias.

\section{Approach}

We developed a trading agent using an integrated deep learning architecture. The core problem was deciding daily whether to buy, sell, or hold a stock to maximize profit based on historical data. Our solution involves three interconnected components processing daily market data formatted as 9x9x12 tensors (representing 9x9 spatial layout across 12 feature channels/time steps):

Convolutional Neural Network (CNN): A seven-layer 2D CNN processes the 9x9x12 input tensor. Each convolutional layer uses LeakyReLU activation. Dropout is applied in the fully connected layers following the convolutional blocks. The primary goal of this CNN component is to identify significant spatial patterns or arrangements within the indicator and price data presented in this grid structure. After processing, the CNN pathway distills its findings into 8 key summary features. These 8 numerical values represent the most important spatial patterns the CNN detected in the input data for that specific trading day.

Long Short-Term Memory (LSTM) with Attention: The same 9x9x12 input tensor is reshaped so the 12 channels become a time sequence of length 12, with each step having 81 features (flattened 9x9). This sequence is fed into a single-layer LSTM with a hidden dimension of 256. An additive Bahdanau attention mechanism is applied to the LSTM's output sequence, allowing the model to weigh the importance of different time steps (channels 0-11, corresponding to specific prices/indicators) when creating a context vector. The LSTM, aided by the attention mechanism, ultimately condenses its analysis into 8 key summary features. These 8 numerical values represent the most significant temporal patterns identified within the sequence of input data.

Deep Q-Network (DQN): The 8 features from the CNN, the 8 features from the LSTM, and a single feature representing the agent's current position (0 for not invested, 1 for invested) are concatenated to form a 17-dimensional state vector. This state vector is the input to the DQN agent, which is a seven-layer fully connected neural network using LeakyReLU activations and dropout. The DQN outputs Q-values for three discrete actions: Hold (0), Sell (1), and Buy (2).

\includegraphics[width=0.8\linewidth]
{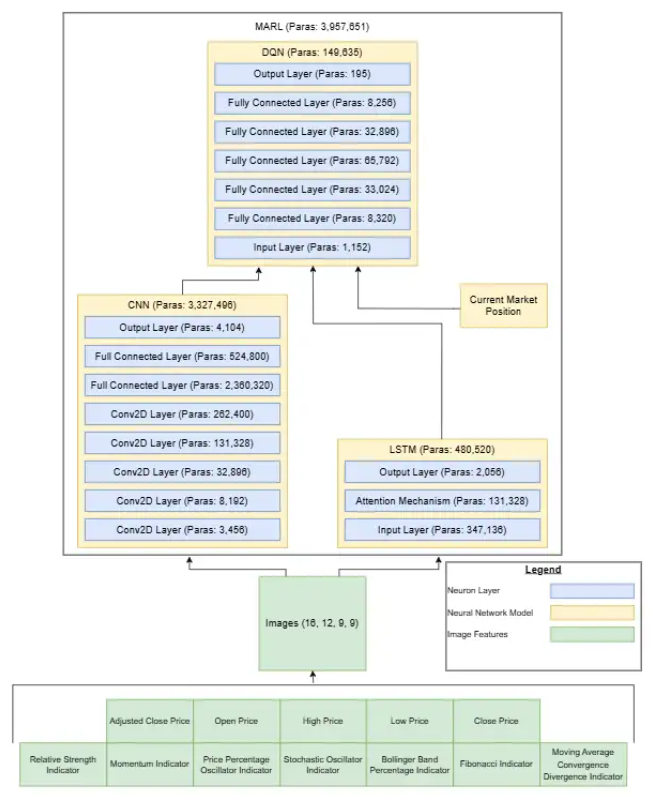}\\
\begingroup\centering
Figure 1: MARL Architecture Diagram\\
\endgroup
The agent learns a trading policy using Q-learning. It steps through the historical training data day by day. At each step, it observes the state, selects an action using an epsilon-greedy strategy (chosen for its simplicity in balancing exploration and exploitation in discrete action spaces), and receives a reward. The reward function is based on the daily percentage change in portfolio value, incorporating penalties for invalid actions (e.g., buying when already holding stock, selling when not holding stock) and estimated costs for commission and market impact. The DQN model is trained using the Mean Squared Error (MSE) loss, a standard choice for Q-value regression, comparing the predicted Q-value for the chosen action against a target Q-value. Crucially, this next-state Q-value is estimated using a separate target network. This target network is initially an exact copy of the main DQN. However, instead of being updated directly by the loss function every step, its weights are updated gradually to track the weights of the main network. In each update step, the target network's weights become a blend of mostly its old weights and a small fraction of the current main network's weights. This process acts like an exponential moving average, ensuring the target network changes smoothly over time rather than copying every rapid fluctuation from the main network, which helps stabilize the learning process.We use the Adam with Decoupled Weight Decay (AdamW) optimizer and a learning rate scheduler. Gradient clipping is applied to prevent exploding gradients. Standard Deep Q-Networks often learn by randomly sampling past experiences from a memory buffer to ensure training stability. However, because financial data is time-dependent, our approach processes the data sequentially, day by day, to preserve the crucial order of events needed for learning valid trading patterns. This sequential processing is necessary because randomly shuffling financial data would break the crucial temporal dependencies needed to learn valid trading patterns.

We believed this approach would succeed by combining the specialized strengths of different neural network architectures. CNNs can identify patterns in the structured spatial encoding of indicators, LSTMs excel at modeling sequences (how indicators evolve over the 12 time steps) [3], and DQNs excel at learning optimal decision-making policies in dynamic environments [2]. By having the CNN and LSTM preprocess the complex market data into a lower-dimensional, more informative state representation, we simplify the task for the DQN agent, allowing it to focus on learning the trading strategy itself [3]. The attention mechanism further enhances the LSTM by allowing it to focus on the most relevant historical indicators (time steps/channels).

While combining CNNs, LSTMs, and RL for finance has been explored [3], our specific integrated architecture, the way data is formatted into a 9x9x12 tensor serving both CNN and LSTM, and the strong emphasis on using a suite of explainability tools (input saliency maps, attention weights, gradient norms, DQN input weights) to analyze and validate the model's decision-making process represent a novel contribution in the context of building trustworthy trading systems.

We anticipated standard reinforcement learning challenges: potential training instability, the difficulty of extracting signals from noisy financial data (addressed by CNN/LSTM feature extraction), and the risk of overfitting (addressed by dropout, AdamW's weight decay, and validation).

\subsection{Problems during development:}
\begin{itemize}
    \item Lack of Transparency: Initially, it was difficult to understand the DQN's behavior – which inputs it prioritized, whether it learned meaningful strategies, and if gradients were flowing correctly. This motivated the development of our visualization suite to track gradient norms, attention weights, input saliency, and feature importance.
    \item Vanishing Gradients: Early experiments showed learning stagnation. By tracking gradient norms, we observed gradients diminishing to near zero in deeper layers or backpropagating weakly to the CNN/LSTM, particularly when using batch normalization across all CNN layers. This led to the removal of batch normalization and adjustments to the architecture.
    \item Model Collapse: At times, visualizations indicated the DQN was effectively ignoring the outputs from the CNN or LSTM, collapsing to a simpler strategy. This often correlates with suboptimal learning rates, requiring careful tuning and the implementation of a learning rate scheduler to ensure balanced learning across components.
    \item Insufficient Model Complexity: Initial, simpler versions of the models struggled to learn effective trading strategies, often failing to make any trades or exhibiting poor performance for many training episodes (e.g., not initiating trades until episode 1600). This indicated the need to increase model capacity (more layers, larger hidden dimensions like the LSTM's 256 units) to capture the complex patterns in the financial data.
    \item Slow Convergence: The lengthy training required (e.g., 1600 episodes for initial trading) raised questions about convergence speed. This might be inherent to the problem's complexity or potentially related to factors like the target network update frequency (`tau`) or the efficiency of the epsilon-greedy exploration strategy in this offline setting.
\end{itemize}

\section{Experiments and Results}

We measured the success of our trading agent through both quantitative performance and qualitative analysis of its internal processes:
\begin{itemize}
    \item Quantitative Metrics: The primary measure was the agent's trading performance on historical data it had not been trained on (the test and validation periods). We simulated the agent's trades day-by-day according to its learned policy and calculated the resulting cumulative portfolio return over the period to assess overall profitability. This was visualized in portfolio value plots.
    \item Qualitative Analysis (Explainability): To understand how the agent made decisions, we employed several visualization techniques:
    \begin{itemize}
        \item Input Feature Importance: Generating visualizations highlighting which parts of the initial input data most strongly influenced the CNN component's analysis.
        \item Temporal Focus: Creating plots showing which specific historical indicators or price types the LSTM component paid the most attention to.
        \item Learned Feature Weights: Visualizing the initial weights within the DQN component to understand the static importance assigned to features from the CNN, LSTM, and investment status.
        \item Learning Dynamics: Tracking the magnitude of learning updates (gradients) flowing back to different parts of the model during training.
    \end{itemize}
\end{itemize}

\includegraphics[width=0.8\linewidth]
{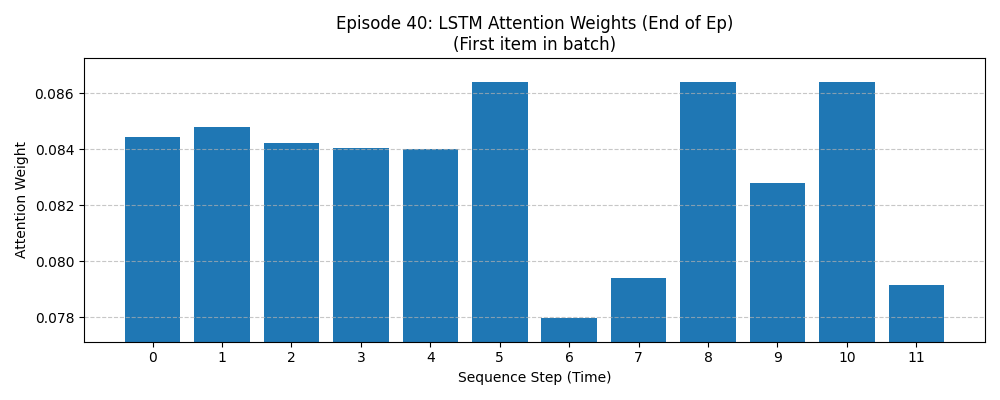}\\
\begingroup\centering
Figure 2: Episode 40 LSTM Attention Weights\\
\endgroup
\includegraphics[width=0.8\linewidth]
{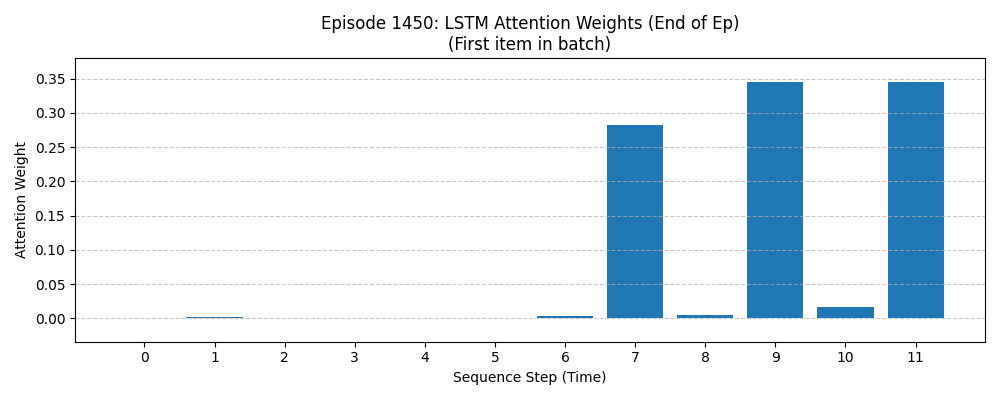}\\
\begingroup\centering
Figure 3: Episode 1450 LSTM Attention Weights\\
\endgroup
The core experiment involved training the integrated CNN-LSTM-DQN agent. The training process involved repeatedly stepping through the historical training data, simulating trading days as distinct episodes. In each step, the agent observed the market state, chose an action, received a calculated reward based on the simulated trade outcome (including estimated costs), and updated its internal neural networks based on this experience.

Periodically during training, we paused the learning process to evaluate the agent's current strategy. This involved running the agent on the training, validation, and separate test datasets, allowing it to make decisions based purely on its learned policy without any random exploration. We calculated the resulting portfolio values for these evaluation periods and generated plots to track performance over time.

Concurrently with these periodic evaluations, we generated the various explainability visualizations described above. This allowed us to monitor the agent's internal decision-making process and learning dynamics as training progressed.

\includegraphics[width=0.8\linewidth]
{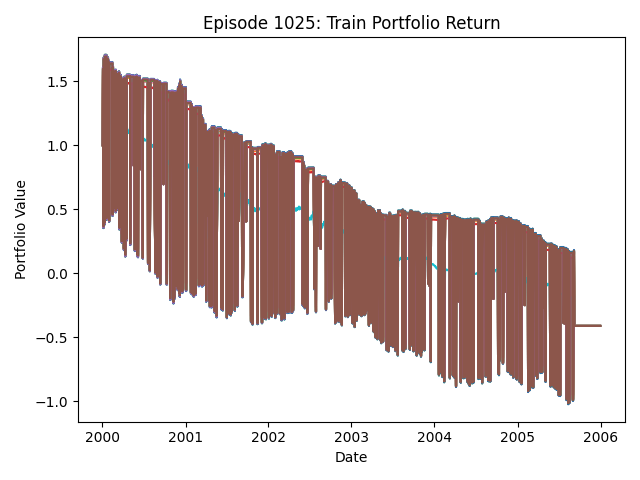}\\
\begingroup\centering
Figure 4: Episode 1025 Training Portfolio Return\\
\endgroup
The agent's performance, particularly on the test set plots, was compared against the performance of a simple passive buy-and-hold strategy applied to the same stocks over the same period. Furthermore, analyzing the visualizations related to input weights and learning signal strength helped assess the relative contribution and interaction of the different model components (CNN, LSTM, DQN).

\includegraphics[width=0.8\linewidth]
{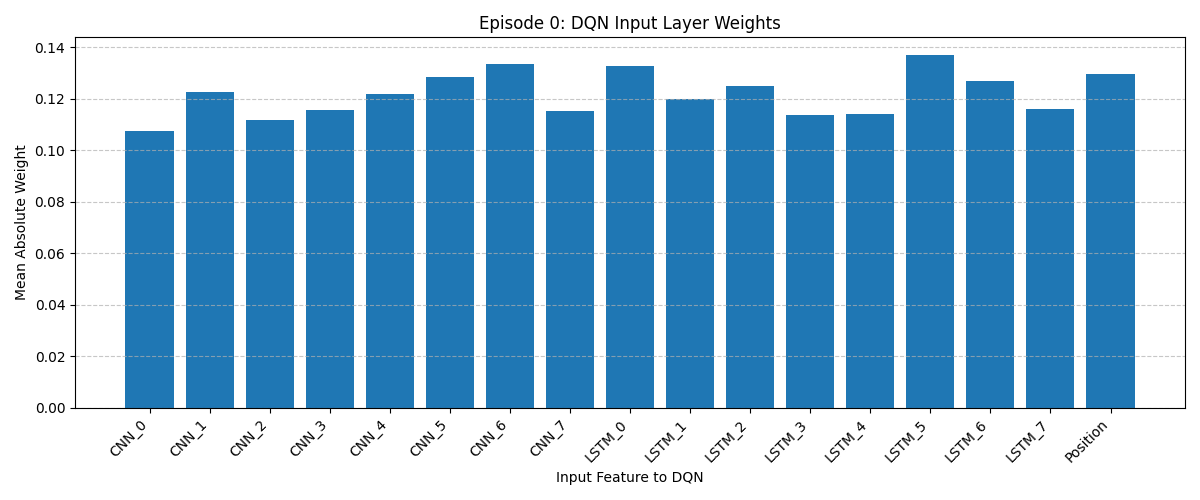}\\
\begingroup\centering
Figure 5: Episode 0 DQN Input Layer Weights\\
\endgroup
\includegraphics[width=0.8\linewidth]
{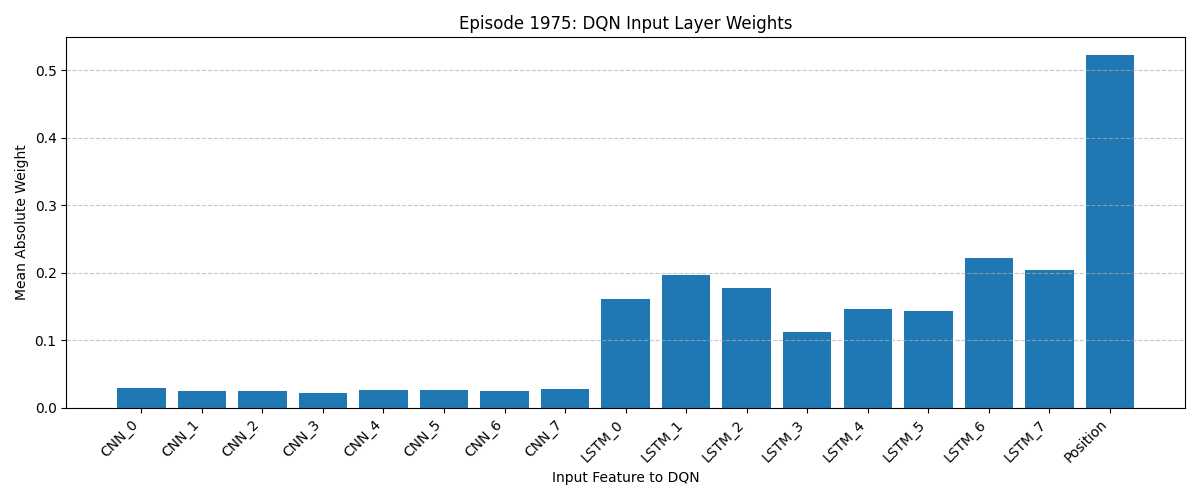}\\
\begingroup\centering
Figure 6: Episode 1975 DQN Input Layer Weights\\
\endgroup
\subsection{Analysis of these results revealed several key findings:}
\begin{itemize}
    \item Performance: The agent's performance on the test set plots allows for a visual comparison against a simple passive buy-and-hold strategy over the same period. These plots illustrate the cumulative returns and volatility resulting from the agent's learned trading decisions, providing a basis for evaluating its effectiveness relative to the benchmark.
    \item Explainability Insights: The input importance visualizations suggested the CNN learned to focus on relevant patterns within the grid-formatted input data, often highlighting areas corresponding to recognizable technical indicator setups.
\end{itemize}

\includegraphics[width=0.8\linewidth]
{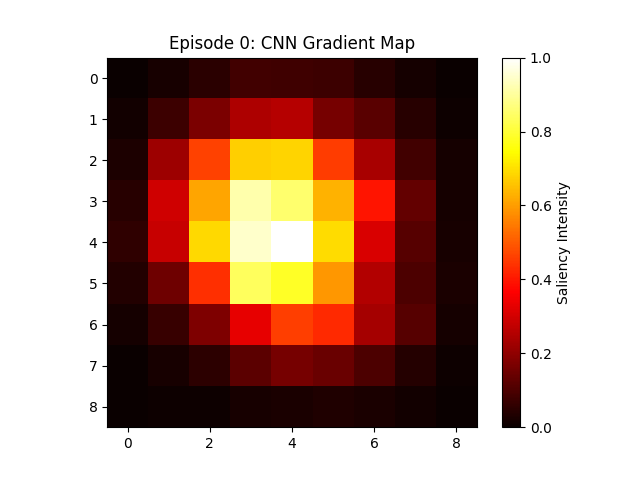}\\
\begingroup\centering
Figure 7: Episode 0 CNN Gradient Map\\
\endgroup
\includegraphics[width=0.8\linewidth]
{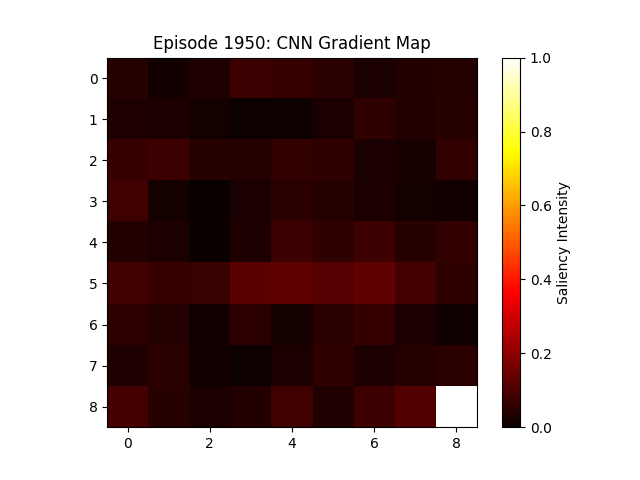}\\
\begingroup\centering
Figure 8: Episode 1950 CNN Gradient Map\\
\endgroup
\begin{itemize}
    \item The LSTM's attention plots showed it dynamically adjusted its focus across the sequence of input indicators and prices, frequently prioritizing more recent data or significant market events, confirming its ability to capture temporal context.
    \item Visualizations of the DQN's initial weights provided insight into how it valued the summary features from the CNN, LSTM, and its own investment status.
    \item Monitoring learning updates confirmed that error signals from the DQN successfully propagated backward to adjust the parameters of the LSTM, attention mechanism, and DQN layers, indicating cohesive learning.
    \item Trading Behavior: Observation of the agent's actions during evaluation suggested it learned a strategy resembling trend-following combined with risk management. It tended to enter positions following favorable internal signals and exit to secure profits or mitigate losses, often choosing to remain inactive during periods of high uncertainty.
\end{itemize}

\includegraphics[width=0.8\linewidth]
{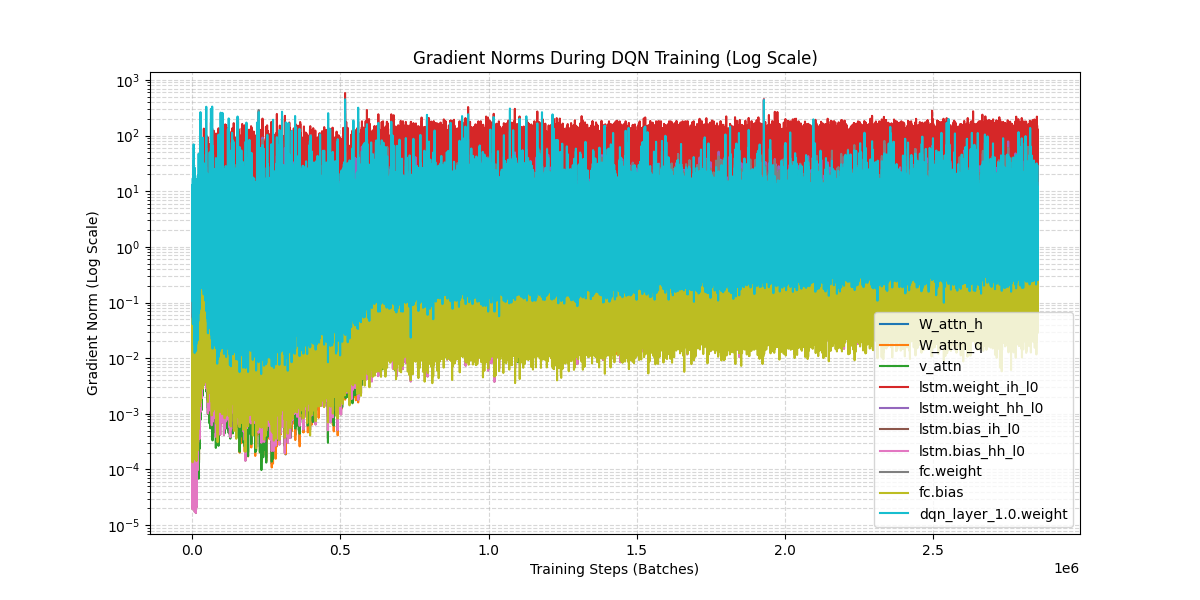}\\
\begingroup\centering
Figure 9: DQN Tracked Gradient Norms\\
\endgroup
Based on the analysis of the model's learning process and internal mechanics, the project succeeded in demonstrating the viability of the integrated CNN-LSTM-DQN architecture. The explainability plots provided evidence that the system was learning, not acting randomly. Specifically, we observed:

\begin{itemize}
    \item Feature Extraction and State Reduction: The CNN and LSTM components successfully processed the input data (prices and indicators) and generated lower-dimensional feature summaries.
    \item Cohesive Learning: Gradient tracking confirmed that learning signals from the DQN's reward-based updates were successfully backpropagated to adjust the weights of the upstream CNN and LSTM components. This indicates the feature extractors were adapting based on the downstream trading task.
    \item Attention and Focus: Visualizations showed the LSTM's attention mechanism dynamically focusing on specific indicators or price types, and saliency maps suggested the CNN identified relevant patterns in the input structure.
    \item DQN Learning: The DQN component demonstrated learning behavior, responding to the reward signals and adjusting its Q-values and resulting policy over training episodes.
\end{itemize}
While any definitive claim about outperforming any market benchmarks would require further validation, these results strongly suggest that the multi-agent architecture effectively integrated information, reduced the state space complexity for the decision-making agent, and demonstrated cohesive learning across its components. This validates the approach as a worthwhile direction for further research in developing sophisticated trading agents.\\

*The project utilized Market Simulation and Technical Indicator code submitted by John Christopher Tidwell and John Storm Tidwell during Machine Learning for Trading 7646 Fall 2024 Semester.

\section{Limitations and Future Work}
While our results are promising, there are several avenues to explore and improvements to consider:
Additional Data Sources: Our current model relies only on technical indicators. Incorporating alternative data could further enhance predictive power. For example, adding a news sentiment analysis component or fundamental indicators could allow the system to anticipate events that pure price data cannot. We envision adding another agent (e.g., a news analysis LSTM or transformer) whose signal could feed into the DQN alongside the CNN and price-LSTM. This multi-agent fusion of technical and fundamental signals might handle market-moving news and events more effectively, reducing vulnerability to sudden shocks.
Advanced RL Algorithms: We used a vanilla DQN for the decision agent. Future work could explore more advanced or tailored RL algorithms: for instance, Double DQN, which addresses overestimation bias, or policy-gradient methods, like Actor-Critic algorithms, which might better handle continuous action spaces or directly optimize risk-adjusted returns. It would be insightful to compare how algorithms like DDPG or PPO perform in this trading setting versus DQN.
Improved Reward Design: We can refine the reward structure to incorporate risk-adjusted returns or other financial metrics. In our current setup, the reward is essentially portfolio profit. In future versions, we could penalize excessive volatility or large drawdowns by using a Sharpe ratio objective or adding a small negative reward for every day the portfolio experiences a loss beyond a threshold. We could also integrate transaction cost penalties to push the agent toward more efficient trading.

\section{Work Division}

\begin{table}[h]
\begingroup\centering
\begin{tabular}{|>{\raggedright\arraybackslash}p{3cm}|>{\raggedright\arraybackslash}p{4cm}|>{\raggedright\arraybackslash}p{7cm}|}
\hline
\textbf{Student Name} & \textbf{Contributed Aspects} & \textbf{Details} \\
\hline
John Storm Tidwell & CNN, DQN and Data Prep & Focused on data preparation, implementing the CNN model and developing the core DQN agent. This included setting up the reinforcement learning environment, defining the reward structure, managing the training loop, and implementing input saliency visualizations. \\
\hline
John Christopher Tidwell & Technical Indicators, LSTM and metrics & Concentrated on implementing technical indicators, building the LSTM and attention mechanism, and evaluating quantitative performance using metrics. Also developed visualizations for LSTM attention weights. Visualizations for layer gradients and DQN weights of the observation vectors. \\
\hline
John Storm Tidwell John Christopher Tidwell & Design, tuning, training and final report & Collaborated on the overall system design, conducted experiments by running and tuning the training process, analyzed the results generated and jointly authored the final report.\\
\hline
\end{tabular}
\caption{Contributions of team members.}
\label{tab:team_contributions}
\endgroup
\end{table}

\subsection{References}

\noindent[1] "Practical Deep Reinforcement Learning Approach for Stock Trading", Xiao-Yang Liu et al.\\[0pt]
[2] "Optimizing Trading Strategies in Quantitative Markets using Multi-Agent Reinforcement Learning", Hengxi Zhang et al.\\[0pt]
[3] "Predicting Stock Market time-series data using CNN-LSTM Neural Network model" , Anurag $M$ Bagde et al.

\end{document}